%% file: naacl2021.tex
\pdfoutput=1

\documentclass[11pt]{article}

\usepackage[]{naacl2021}

\usepackage{times}
\usepackage{latexsym}

\usepackage[T1]{fontenc}

\usepackage[utf8]{inputenc}

\usepackage{microtype}

\usepackage{graphicx}               %
\usepackage{tabularx}               %
\newcolumntype{C}{>{\centering\arraybackslash}X}
\usepackage{multirow}               %
\usepackage{diagbox}                %
\usepackage{hhline}                 %
\usepackage{color}                  %
\usepackage{amsmath}                %
\usepackage{amssymb}                %
\usepackage{mathtools}              %
\usepackage{enumitem}               %

\usepackage{subfigure}
\usepackage{booktabs}
\usepackage{extarrows}
\usepackage{makecell}
\usepackage{multirow}
\usepackage[ruled]{algorithm2e}

\newtheorem{thm:def}{Definition}
\newtheorem{thm:eg}{Example}
\newtheorem{thm:lem}{Lemma}

\definecolor{RoseQuartzBg}{HTML}{F7CAC9}
\definecolor{RoseQuartz}{HTML}{F5A798}
\definecolor{Serenity}{HTML}{92A8D1}
\definecolor{OrangeRed}{rgb}{1.0, 0.27, 0.0}
\definecolor{Turquoise}{HTML}{0F4C81}
\usepackage{xparse}

\definecolor{pigment}{rgb}{0.2, 0.2, 0.6}

\newcommand{\ours}{BART-Gen}

\title{Document-Level Event Argument Extraction by Conditional Generation }

\author{Sha Li \and Heng Ji \and Jiawei Han \\
 University of Illinois at Urbana-Champaign, IL, USA \\
 \texttt{\{shal2, hengji, hanj\}}@illinois.edu
}

\begin{document}
\maketitle

\begin{abstract}
\input{abs}

\end{abstract}

\input{intro}

\input{method}

\input{dataset}

\input{exp}

\input{related}

\input{conclusion}

\section*{Acknowledgement}
The authors would like to thank Tuan Lai and Zhenhailong Wang for their help in the data processing and annotation organization effort.
We also thank all the annotators who have contributed to the annotations of our training data (in alphabetical order): Daniel Campos, Anthony Cuff, Yi R. Fung, Xiaodan Hu, Emma Bonnette Hamel, Samual Kriman, Meha Goyal Kumar, Manling Li, Tuan M. Lai, Ying Lin, Sarah Moeller, Ashley Nobi, Xiaoman Pan, Nikolaus Parulian, Adams Pollins, Rachel Rosset, Haoyu Wang, Qingyun Wang, Zhenhailong Wang, Spencer Whitehead, Lucia Yao, Pengfei Yu, Qi Zeng, Haoran Zhang, Hongming Zhang, Zixuan Zhang.

This research is based upon work supported by U.S. DARPA KAIROS Program No. FA8750-19-2-1004, U.S. DARPA AIDA Program No. FA8750-18-2-0014, National Science Foundation IIS-19-56151, IIS-17-41317, and IIS 17-04532, and Air Force No. FA8650-17-C-7715. The views and conclusions contained herein are those of the authors and should not be interpreted as necessarily representing the official policies, either expressed or implied, of DARPA, or the U.S. Government. The U.S. Government is authorized to reproduce and distribute reprints for governmental purposes notwithstanding any copyright annotation therein.

\bibliography{anthology,ref}
\bibliographystyle{acl_natbib}

\end{document}

%% file: abs.tex
Event extraction has long been treated as a sentence-level task in the IE community. We argue that this setting does not match human information seeking behavior and leads to incomplete and uninformative extraction results.
We propose a document-level neural event argument extraction model by formulating the task as conditional generation following event templates. We also compile a new document-level event extraction benchmark dataset \textsc{WikiEvents} which includes complete event and coreference annotation. On the task of argument extraction, we achieve an absolute gain of 7.6\% F1 and 5.7\% F1 over the next best model on the \textsc{RAMS} and \textsc{WikiEvents} datasets respectively. On the more challenging task of informative argument extraction, which requires implicit coreference reasoning, we achieve a 9.3\% F1 gain over the best baseline.  
To demonstrate the portability of our model, we also create the first end-to-end zero-shot event extraction framework and achieve 97\% of 
fully supervised model's trigger extraction performance and 82\% of the argument extraction  performance given only access to 10 out of the 33 types on \textsc{ACE}.
\footnote{The programs, data and resources are publicly available for research purpose at \url{https://github.com/raspberryice/gen-arg}.}

%% file: intro.tex
\section{Introduction}

By converting a large amount of unstructured text into trigger-argument structures, event extraction models provide unique value in assisting us process volumes of documents to form insights. While real-world events are often described throughout a news document (or even span multiple documents), the scope of operation for existing event extraction models have long been limited to the sentence level.

Early work on event extraction originally posed the task as document level role filling~\cite{Grishman1996MUC6} 
on a set of narrow scenarios %
and evaluated on small datasets. 
The release of ACE\footnote{\url{https://www.ldc.upenn.edu/collaborations/past-projects/ace}}, a large scale dataset with complete event annotation, opened the possibility of applying powerful machine learning models 
which led to substantial improvement in event extraction. The success of such models and the widespread adoption of ACE as the training dataset established sentence-level event extraction as the mainstream task defintion. 

\begin{figure}
    \centering
    \includegraphics[width=\linewidth]{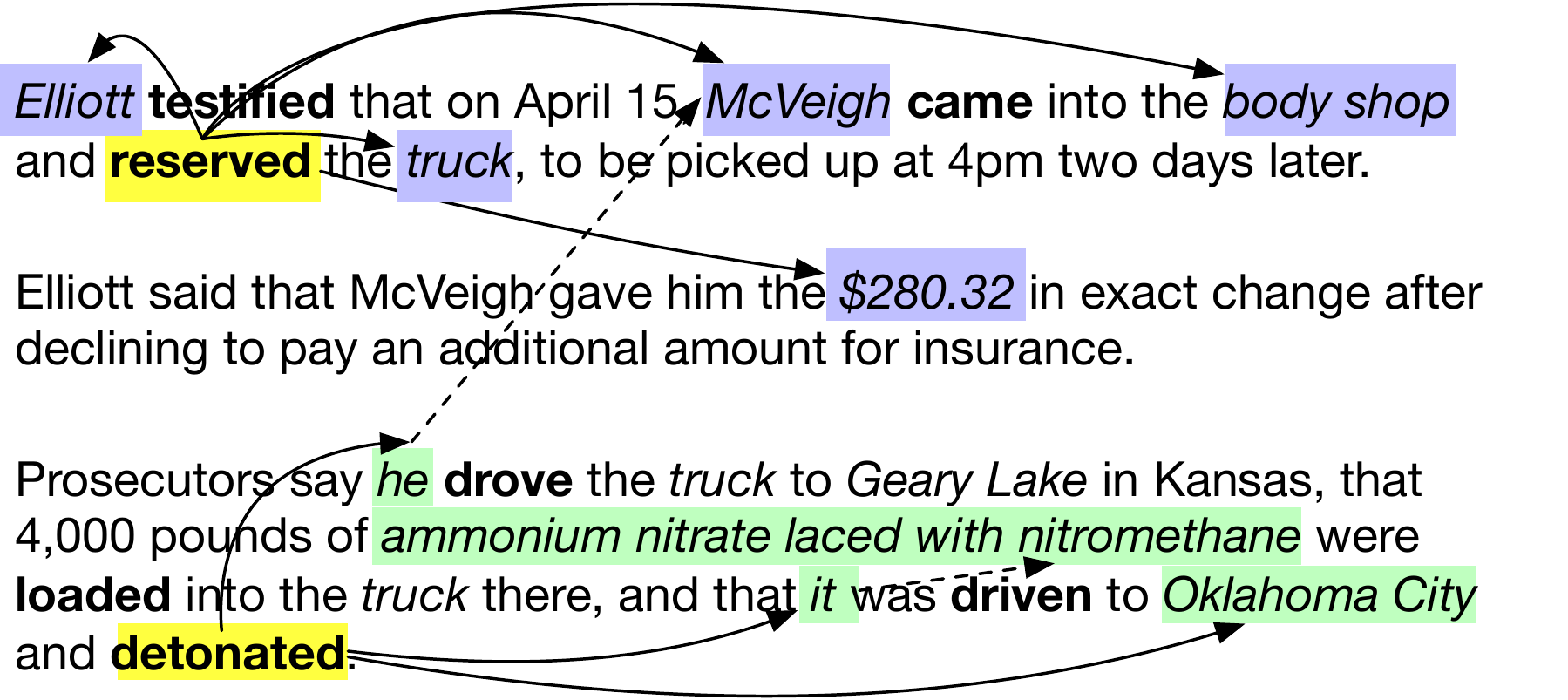}
    \caption{Two examples of cross-sentence inference for argument extraction from our \textsc{WikiEvents} dataset. The PaymentBarter argument of the Transaction.ExchangeBuySell event triggered by ``reserved'' in the first sentence can only be found in the next sentence. 
    The Attack.ExplodeDetonate event triggered by ``detonated'' in the third sentence has an uninformative argument ``he'', which needs to be resolved to the name mention ``McVeigh'' in the previous sentences. 
    }
    \label{fig:intro-ex}
     \vspace{-0.4cm}
\end{figure}
This formulation signifies a misalignment between the information seeking behavior in real life and the exhaustive annotation process in creating the datasets.
An information seeking session~\cite{mai2016looking} can be divided into 6 stages: task initiation, topic selection, pre-focus exploration, focus information, information collection and search closure~\cite{kuhlthau1991searchprocess}. Given a target  event ontology, we can safely assume that topic selection is complete and users start from skimming the documents before they discover events of interest, focus on such events and then aggregate all relevant information for the events. In both the ``pre-focus exploration'' and ``information collection'' stages, users naturally cross sentence boundaries. 

Empirically, using sentence boundaries as event scopes conveniently simplifies the problem, but also introduces fundamental flaws: the resulting extractions are \textit{incomplete} and \textit{uninformative}.
We show two examples of this phenomenon in Figure \ref{fig:intro-ex}. 
The first example exemplifies the case of implicit arguments across sentences. 
The sentence that contains the PaymentBarter argument ``\$280.32" is not the sentence that contains the trigger ``reserve" for the ExchangeBuySell event.
Without a document-level model, such arguments would be missed and result in \textit{incomplete} extraction.
In the second example, the arguments are present in the same sentence, but written as pronouns. Such extraction would be \textit{uninformative} to the reader without cross-sentence coreference resolution.

\begin{figure*}[t]
    \centering
    \includegraphics[width=\linewidth]{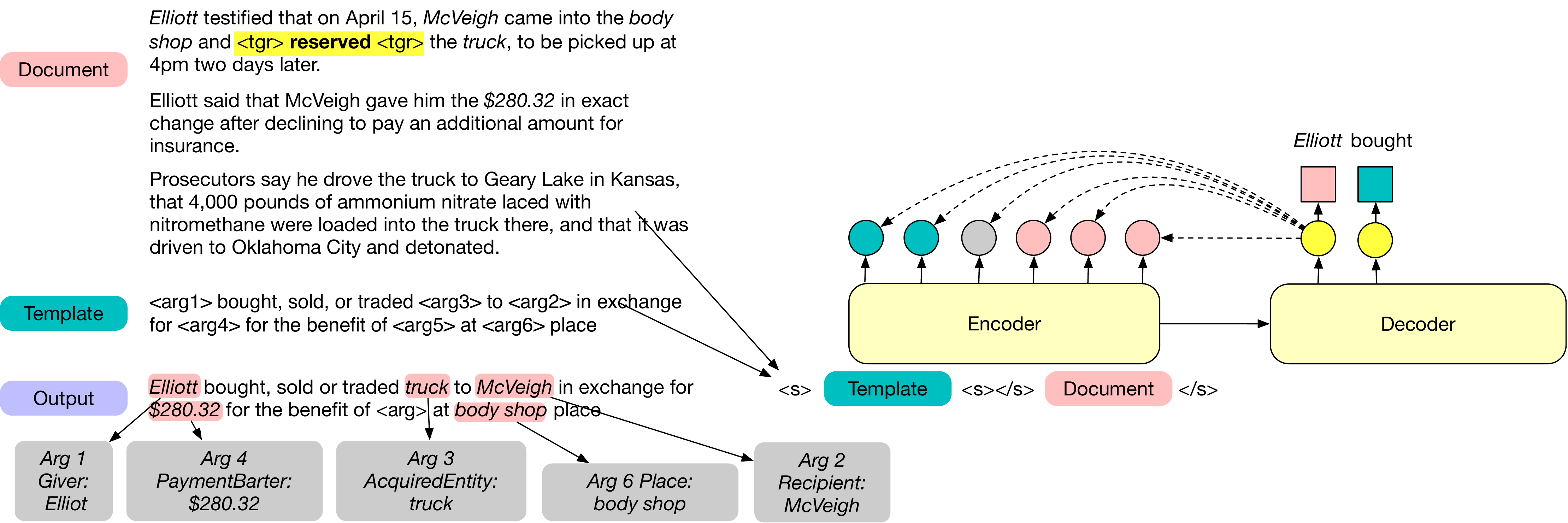}
    \caption{Our argument extraction model using conditional generation. On the left we show an example document, template and the desired output for the instance.  Each example document may contain multiple event triggers and we use special $\langle tgr \rangle$ tokens to markup the target event trigger for argument extraction (the highlighted word ``reserved"). The input to the model is the concatenation of the template and the document. The decoded tokens are either from the template or the document. The color of the generated tokens indicate its copy source. After the filled template is generated, we extract the spans to produce the final output. 
   }
    \label{fig:model}
  \vspace{-0.3cm}
\end{figure*}

We propose a new end-to-end document-level event argument extraction model by framing the problem as \textit{conditional generation} given a template. Conditioned on the unfilled template %
and a given context, the model is asked to generate a filled-in template with arguments as shown in Figure \ref{fig:model}. Our model does not require entity recognition nor coreference resolution as a preprocessing step and can work with long contexts beyond single sentences.
Since templates are usually provided as part of the event ontology definition, this requires no additional human effort. 
Compared to recent efforts~\cite{Du2020EventQA, Feng2020ProbingAF, Chen2019ReadingTheManual} that retarget question answering (QA)  models for event extraction, our generation-based model can easily handle the case of missing arguments and multiple arguments in the same role without the need of tuning thresholds and can extract all arguments in a single pass. 

In order to evaluate the performance of document-level event extraction, we collect and annotate a new benchmark dataset \textsc{WikiEvents}.
This document-level evaluation also allows us to move beyond the nearest mention of the argument and instead seek the most informative mention\footnote{We prefer name mentions over nominal mentions and only use pronoun mentions when no other mentions exist.} in the entire document context. 
In particular, only 34.5\% of the arguments detected in the same sentence as the trigger can be considered informative.  We present this new task of \textit{document-level informative argument extraction} and show that while this task requires much more cross-sentence inference, our model can still perform reliably well. 

Since we provide the ontology information (which roles are needed for the event) through the template as an external condition, our model has excellent portability to unseen event types.  By pairing up our argument extraction model with a keyword-based zero-shot trigger extraction model, we enable zero-shot transfer for new event types.

The major contributions of this paper can be summarized as follows:

\begin{enumerate}
    \item We address the  document-level argument extraction task with an end-to-end neural event argument extraction model by conditional text generation. Our model does not rely on entity extraction nor entity/event coreference resolution. Compared to QA-based approaches, it can easily handle missing arguments and multiple arguments in the same role.
    \item We present the first \textit{document-level event extraction} benchmark dataset with \textit{complete event and coreference annotation}. We also introduce the new \textit{document-level informative argument extraction} task, which evaluates the ability of models to learn entity-event relations over long ranges.
    \item We release the first end-to-end zero-shot event extraction framework by combining our argument extraction model with a zero-shot event trigger classification model. %
\end{enumerate}

%% file: method.tex
\section{Method}
\vspace{-0.2cm}
The event extraction task consists of two subtasks: trigger extraction and argument extraction. 
The set of possible event types and roles for each event type are given by the event ontology as part of the dataset. One  template for each event type is usually pre-defined in the ontology. \footnote{ACE does not come with templates, but since the event types are subsets of the RAMS AIDA ontology and the KAIROS ontology, we reused templates from the these ontologies.}

We first introduce our document-level argument extraction model in Section 2.1 and then introduce our zero-shot keyword-based trigger extraction model in Section 2.2.

\subsection{Argument Extraction Model}

We use a conditional generation model for argument extraction, where the condition is an unfilled template and a context. The template is a sentence that describes the event with $\langle arg \rangle$ placeholders. The generated output is a filled template where placeholders are replaced by concrete arguments. An example of the unfilled template from the ontology and the filled template for the event type Transaction.ExchangeBuySell \footnote{This type is used for a transaction transferring or obtaining money, ownership, possession, or control of something, applicable to any type, nature, or method of acquisition including barter.} can be seen in Figure \ref{fig:model}. Notably, one template per event type is given in the ontology, and does not require further human curation as opposed to the question designing process in question answering (QA) models~\cite{Du2020EventQA, Feng2020ProbingAF}.

Our base model is an encoder-decoder language model (BART~\cite{Lewis2020BARTDS}, T5~\cite{Raffel2020T5}. The generation process models the conditional probability of selecting a new token given the previous tokens and the input to the encoder. 
\begin{equation}
\small 
p(x \mid c)=\prod_{i=1}^{|x|} p\left(x_{i} \mid x_{<i}, c\right)
\vspace{-0.4em}
\end{equation}
In the encoder, bidirectional attention layers are used to enable interaction between every pair of tokens and produce the encoding for the context $c$.
Each layer of the decoder performs cross-attention over
the output of the encoder in addition to the attention over the previous decoded tokens.

To utilize the encoder-decoder LM for argument extraction, we construct an input sequence of 
$\langle$s$\rangle$ template $\langle s\rangle \langle / s \rangle$document $ \langle / s \rangle$.
All argument names (arg1, arg2, etc.) in the template are replaced by a special placeholder token $\langle$arg$\rangle$. 
The ground truth sequence is the filled template where the placeholder token is replaced by the argument span whenever possible.
In the case where there are multiple arguments for the same slot, we connect the arguments with the word ``and".

The generation probability is computed by taking the dot product between the decoder output and the embeddings of tokens from the input.

\begin{equation}
\resizebox{\hsize}{!}{
$p(x_i=w | x_{< i}, c,t ) =  
\begin{cases}
\text{Softmax}\left( h_i^T \text{Emb}(w) \right)  & w \in V_c  \\
 0  & w \notin V_c 
\end{cases}$
}
\end{equation}

To prevent the model from hallucinating arguments, we restrict the vocabulary of words to $V_c$: the set of tokens in the input.

The model is trained by minimizing the negative loglikelihood over all (content, template, output) instances in the dataset $D$:
\begin{equation}
\small 
\mathcal{L}(D)=-\sum_{i=1}^{|D|} \log p_{\theta}\left(x^i \mid c^{i} \right)
\end{equation}

\begin{table*}[t]
    \centering
    \small
    \begin{tabular}{m{23em}|m{12em}| m{10em} }
    \toprule 
    Context & Original  & After \\
    \midrule 
    When outlining her tax reform policy , Clinton has made clear that she wants to tax the wealthy and make sure they ``pay their fair share ." \textbf{She} has \underline{proposed} (PublicStatement) a \textbf{tax plan} that would require millionaires and billionaires to pay more taxes than middle-class and lower -income individuals.     &  \textbf{She} communicated with \textbf{tax plan}  about  $\langle$ arg $\rangle$ at  $\langle$ arg $\rangle$  place. {\color{red}\textbf{She } is a person/organization/country. \textbf{tax plan} is a person/organization/country.} &  \textbf{She} communicated with $\langle$ arg $\rangle$ about \textbf{tax plan} at $\langle$ arg $\rangle$ place. {\color{red}\textbf{She } is a person/organization/country. } \\
    \bottomrule 
    \end{tabular}
    \caption{Example of adding type constraints through clarification. The Participant argument of PublicStatement event can only be a person, organization or geo-political entity. The Topic argument can be any type of entity.}
    \label{tab:type-example}
\end{table*}
The event ontology often imposes entity type constraints on the arguments. When using the template only, the model has no access to such constraints and can generate seemingly fluent and sensible responses with the wrong arguments. Inspired by \cite{shwartz2020unsupervised}, we use clarification statements to add back constraints without breaking the end-to-end property of the model.

In the example presented in Table \ref{tab:type-example}, we can see that the greedy decoding selects ``tax plan" as the second Participant argument for the PublicStatement event.
Apart from the preposition ``with", there is nothing in the template indicating that this slot should be filled in with a person instead of a topic. 
To remedy this mistake, we append ``clarifications" for its argument fillers in the form of type statements: $\langle$ arg $\rangle$ is a $\langle$ type $\rangle$. 
We then rerank the candidate outputs by the language modeling probability of the filled template and clarifications. When there are multiple valid types, we take the maximum probability of the valid type statements.
\begin{equation}
\small 
    \log p(x| c) = \sum_i \log p(x_i | x_{<i}, c) + \max_{e \in E_r} \log p(z_e | x, c ) 
\end{equation}
$E_r$ is the set of valid entity types for the role $r$ according to the ontology and $z_e$ is the type statement. 
Since ``tax plan is a person." goes against commonsense, the probability of generating this sentence will be low.  
In this way, we can prune responses with conflicting entity types.

\subsection{Keyword-Based Trigger Extraction Model}
Our argument extraction model relies on detected event triggers (type and offset) as input. Any trigger extraction model could be used in practice, but here we describe a trigger extraction model designed to work with only keyword-level supervision. For example for the ``StartPosition" event, we use 3 keywords ``hire, employ and appoint" as initial supervision with no mention level annotation. \footnote{In a fully supervised setting, it might be desirable to use a supervised trigger extraction model for optimal performance.} This module allows quick transfer to new event types of interest.

We treat the trigger extraction task as sequence labeling and 
our model is an adaptation of TapNet~\cite{Yoon2019TapNetNN, Hou2020FewshotST}, which was designed for few-shot classification and later extended to Conditional Random Field (CRF) models. Compared with \cite{Hou2020FewshotST}, we do not collapse the entries of the transition matrix, making it possible for our model to learn different probabilities for each event type.
Since our model takes class keywords as input, we refer to this model as \textsc{TapKey}. %

For each event type, we first obtain a class representation vector $c_k$ based on given keywords using the masked category prediction method in \cite{Meng2020TextCU}. 
This class representation vector is an average over the BERT vector representations of the keywords, with some filtering applied to remove ambiguous occurrences. 
Details of the filtering process are included in Appendix A.

Following the linear-chain CRF model, the probability of a tagged sequence is:
\begin{equation}
\small 
   \log  p(y | h ; \theta) \propto  \sum_i  \varphi(y_i|h_i)  + \sum_i \psi (y_i | y_{i-1}, h_i)
\end{equation}
$h_i$ is the output of the embedding network (in our case, BERT-large) corresponding to $x_i$.

The label space for $y_i$ is the set of IO tags. We choose to use this simplified tagging scheme because it has fewer parameters and the fact that consecutive triggers of the same event are very rare.

The feature function $\varphi(\cdot)$ is defined as 
\begin{equation}
    \varphi (y_i=k | h_i) = \text{Softmax}  \left(M(h_i)^T M(\phi_k)\right)
\end{equation}
$\phi_k$ is a normalized  reference vector for class $k$ and $M$ is a projection matrix, both of which are parameters of the model.
$M$ is not a learned parameter, but solved by taking the QR decomposition of a modified reference vector matrix. Specifically, $M$ satisfies the following equation: 
\begin{equation}
    M^T (c_k - \lambda \hat \phi_k) = 0
\end{equation}

We refer to the TapNet~\cite{Yoon2019TapNetNN} paper for details and also provide a simplified derivation in Appendix A.

The transition score $\psi(\cdot)$ between tags is parameterized using two diagonal matrices $W$ and $W_o$:
\begin{equation}
\resizebox{0.9\hsize}{!}{
	$
   \psi (y_i=k| y_{i-1}=l, h_i) = \left\{
\begin{array}{l c}
    M(\phi_k) W M(h_i)  &   k = l\neq 0 \\
    M(\phi_k) W_o M(h_i)&  k \text{ or } l =0\\
    0 & k\neq l
\end{array}
\right.$
}
\end{equation}

In the training stage, the model parameters $\{\phi, W, \theta_f \}$ are learned by minimizing the negative log probability of the sequences.
\begin{equation}
\small 
    \mathcal{L} =  -  \frac{1}{N}\sum \log p(y| h; \theta) + \alpha \Vert \Phi^T \Phi - I \Vert^2   
\end{equation}
The matrix $\Phi$ of all reference vectors is initialized as a diagonal matrix and the second term regularizes the vectors to be close to orthonormal during training. $\alpha$ is a hyperparameter. 

In the zero-shot setting, we first train on pseudo labeled data before we apply the model. In the pseudo labeling stage, we directly use the cosine similarity between class vectors and the embeddings of the tokens from the language model to assign labels to text. We only use labels with high confident for both event I tags and O tags. The remainder of the tokens will be tagged as X for unknown.
Then we train the model on the token classification task.  Since none of the parameters in the model are class-specific, the model can be used in a zero-shot transfer setting.

%% file: dataset.tex
\section{Benchmark Dataset \textsc{WikiEvents}}
\begin{table}[t]
    \centering
    \small 
    \begin{tabular}{l|c c c }
    \toprule 
         & Train & Dev & Test  \\
        \midrule 
         \# Event types &  49 & 35 & 34 \\
         \# Arg types & 57 & 32 & 44 \\
         \midrule 
        \# Docs  & 206 & 20 & 20 \\
        \# Sentences &  5262 & 378 & 492 \\
        \# Events & 3241 & 345 & 365 \\
        \bottomrule
    \end{tabular}
    \caption{Statistics for the \textsc{WikiEvents} dataset.}
    \vspace{-0.2cm}
    \label{tab:kairos-stats}
\end{table}
\begin{figure}
    \centering
    \includegraphics[width=\linewidth]{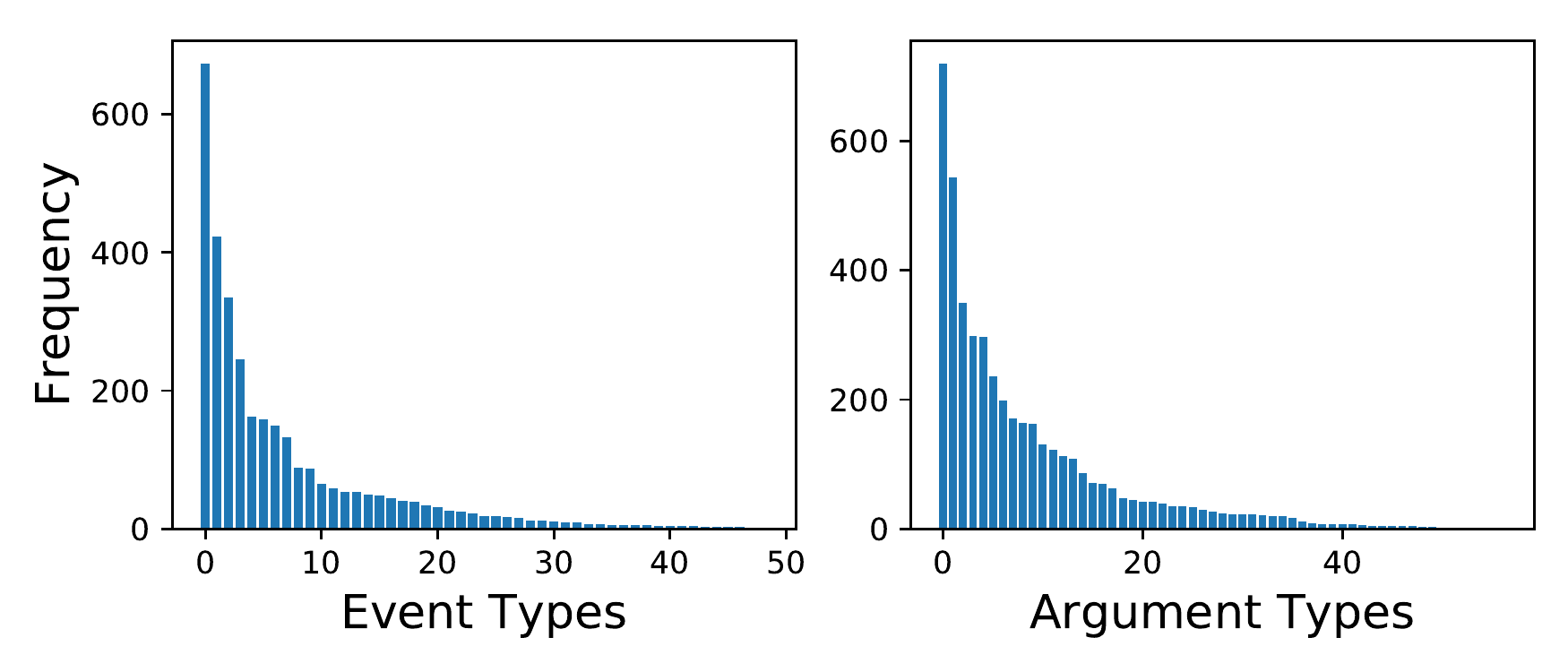}
    \caption{Distribution of event types and argument types in the \textsc{WikiEvents} dataset. }
    \label{fig:kairos_dist}
\end{figure}

\begin{figure}
    \centering
    \includegraphics[width=0.9\linewidth]{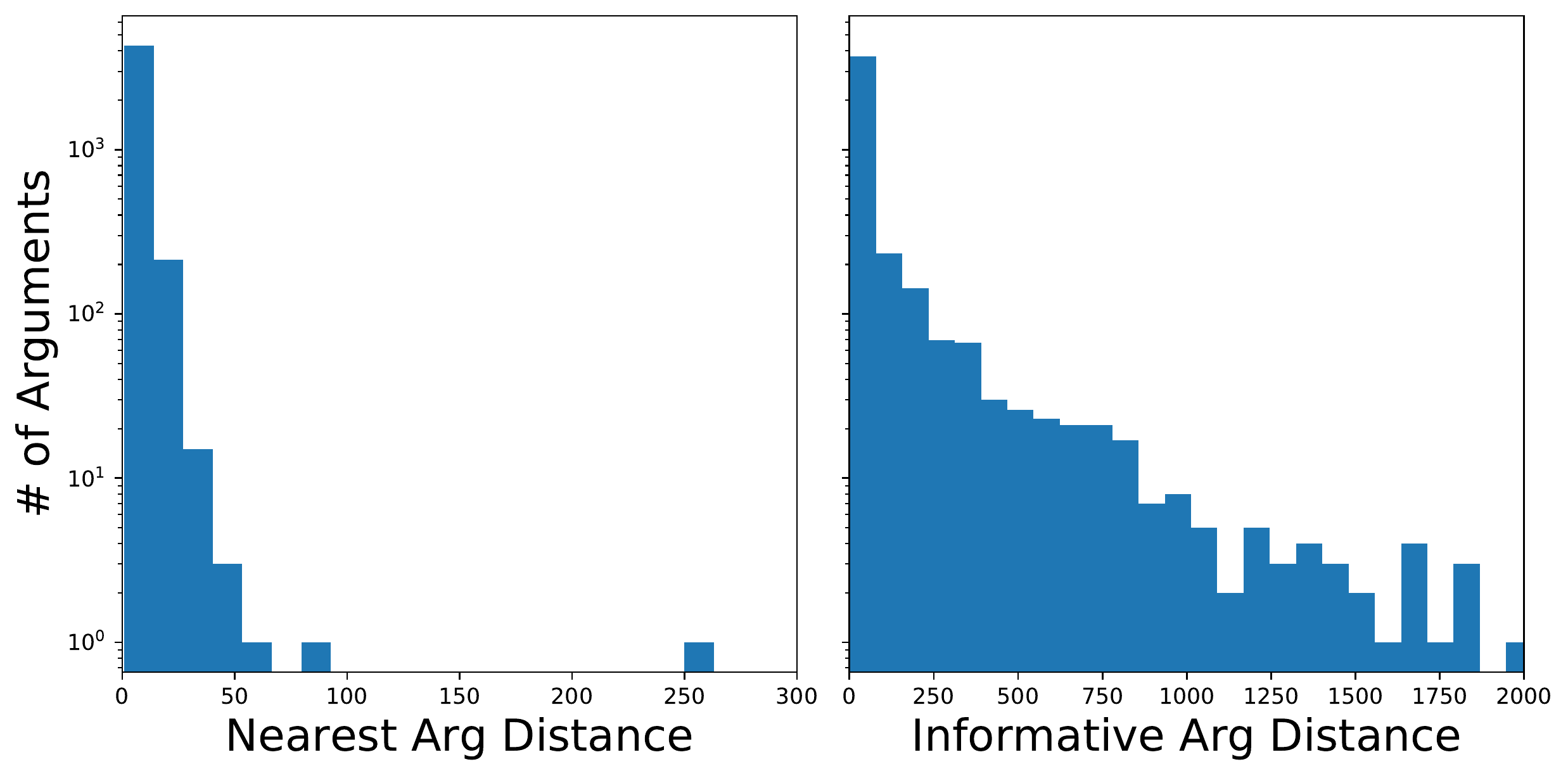}
    \caption{Distribution of distance between event trigger and arguments. Distance is measured in number of words.}
     \vspace{-0.2cm}
    \label{fig:kairos_arg_dist}
\end{figure}

\subsection{Evaluation Tasks}
Our dataset evaluates two tasks: \textit{argument extraction} and \textit{informative argument extraction}. 

For \textit{argument extraction}, we use head word F1 (Head F1) and coreferential mention F1 (Coref F1) as metrics. 
We consider an argument span to be correctly identified if the offsets match the reference. If the argument role also matches, we consider the argument is correctly classified. Since annotators are asked to annotate the head word when possible,
we refer to this metric as Head F1.
For Coref F1, the model is given full credit if the extracted argument is coreferential with the gold-standard argument as used in \cite{Ji2008RefiningEE}.

For downstream applications such as knowledge base construction and question answering, argument fillers that are pronouns will not be useful to the user. Running an additional coreference resolution model to resolve them will inevitably introduce propagation errors. 
Hence, we propose a new task: \textit{document-level informative argument extraction}. We define name mentions to be more informative than nominal mentions, and pronouns to be the least informative. When the mention type is the same, we select the longest mention as the most informative one.
Under this task, the model will only be given credit if the extracted argument is the most informative mention in the entire document. %

\subsection{Dataset Creation}
We collect English Wikipedia articles that describe real world events and then follow the reference links to crawl related news articles.
We first manually identify category pages such as \url{https://en.wikipedia.org/wiki/Category:Improvised_explosive_device_bombings_in_the_United_States} and then for each event page (i.e. \url{https://en.wikipedia.org/wiki/Boston_Marathon_bombing}), we record all the links in its ``Reference" section and use an article scraping tool\footnote{\url{https://github.com/codelucas/newspaper}} to extract the full text of the webpage. 

We follow the recently established ontology from the KAIROS project\footnote{https://www.ldc.upenn.edu/collaborations/current-projects} for event annotation.
This ontology defines 67 event types in a three level hierarchy. In comparison, the commonly used ACE ontology has 33 event types defined in two levels.

We hired graduate students as annotators and provided example sentences for uncommon event types. 
A total of 26 annotators were involved in the process. We used the BRAT\footnote{\url{https://brat.nlplab.org/}} interface for online annotation.

The annotation process is divided into 2 stages: event mention (trigger and argument) annotation and event coreference annotation. In addition to coreferential mention clusters, we also provide the most informative mention for each cluster.
Details about the data collection and annotation process can be found in Appendix B.

\subsection{Dataset Analysis}
Overall statistics of the dataset are listed in Table \ref{tab:kairos-stats}.
Compared to \textsc{ACE}, our \textsc{WikiEvents} dataset has a much richer event ontology, especially for argument roles. 
The observed distributions of event types and argument roles are shown in Figure \ref{fig:kairos_dist}.

We further examine the distance between the event trigger and arguments in Figure \ref{fig:kairos_arg_dist}. When considering the nearest argument mention, the distribution of the arguments is very concentrated towards 0, showing that this annotation standard favors local extractions.
In the case of extracting informative mentions, we have a relatively flat long tail distribution with the average distance being 68.82 words (compared to 4.75 words for the nearest mention). 
In particular, only 34.5\% of the arguments detected in the same sentence as the trigger can be considered informative.
This confirms the need for document level inference in the search of informative argument fillers.

%% file: exp.tex
\section{Experiments}
Our experiments fall under three settings: (1) document-level event argument extraction; (2) document-level informative argument extraction and (3) zero-shot event extraction. 

For document-level event argument extraction we follow the conventional approach of regarding the argument mention with closest proximity to the trigger as the ground truth. In the second setting we consider the most informative mention of the argument as the ground truth.

The zero-shot setting examines the portability of the model to new event types. Under this setting we consider a portion of the event types to be known and only annotation for these event types will be seen. We used two settings for selecting known types: 10 most frequent events types and 8 event types, one from each parent type of the event ontology. The evaluation is done on the complete set of event types.
We refer the reader to Appendix C for implementation details and hyperparameter settings.

\subsection{Datasets}
In addition to our dataset \textsc{WikiEvents}, we also report the performance on the Automatic Content Extraction (\textsc{ACE}) 2005 dataset\footnote{\url{https://www.ldc.upenn.edu/collaborations/past-projects/ace}} and the Roles Across Multiple Sentences (\textsc{RAMS}) dataset\footnote{\url{http://nlp.jhu.edu/rams}}. 

We follow preprocessing from \cite{Lin2020OneIE, Wadden2019EntityRA} for the \textsc{ACE} dataset.
\footnote{Note that our preprocessing procedure is slightly different from \cite{Du2020EventQA} as we kept pronouns as valid event triggers and arguments.}
Statistics of the \textsc{ACE} data splits can be found in Table \ref{tab:ace-stats}.
\begin{table}[]
    \centering
    \small 
    \begin{tabular}{l| c c c  }
    \toprule 
        Split & Event Types & \# Sents & \# Events \\
        \midrule 
        \textbf{Full Training} & 33  & 17172 & 4202  \\
        \textbf{Freq} & 10 & 17172 & 3398  \\
        \textbf{Ontology} & 8 & 17172 & 1311  \\
        \textbf{Dev} & - & 923 &   450  \\
        \textbf{Test} & - & 832 &   403  \\
        \bottomrule
    \end{tabular}
    \caption{Dataset statistics for \textsc{ACE} under multiple settings. In the \textbf{Freq} split, we keep the 10 most frequent event types. In the \textbf{Ontology} split, we keep 1 event subtype per general type in LIFE, MOVEMENT, TRANSACTION, BUSINESS, CONFLICT, CONTACT, PERSONNEL and JUSTICE. \footnote{The selected types are 'Movement:Transport', 'Personnel:Elect',
'Business:Start-Org', 'Life:Injure', 'Transaction:Transfer-Money',
'Justice:Arrest-Jail', 
'Contact:Phone-Write', 'Conflict:Demonstrate'.} }
    \label{tab:ace-stats}
\end{table}
\textsc{RAMS}~\cite{Ebner2020RAMS} is a recently released dataset with cross-sentence argument annotation. A 5-sentence window is provided for each event trigger and the closest argument span is annotated for each role. 
We follow the official data splits from Version 1.0.

\subsection{Document-Level Event Argument Extraction}
Table \ref{tab:RAMS-AE} shows the performance for argument extraction on \textsc{RAMS}. 
On the \textsc{RAMS} dataset, we mainly compare with Two-step~\cite{Zhang2020TwoStep}, which is the current SOTA on this dataset.  To handle long contexts, it breaks down the argument extraction into two steps: head detection and expansion.  

In Table \ref{tab:KAIROS-AE} we show the results for the \textsc{WikiEvents} dataset. We compare with a popularly used BERT-CRF baseline~\cite{shi2019simpleBERTCRF} that performs trigger extraction on sentence-level and BERT-QA~\cite{Du2020EventQA} ran on sentence-level and document-level.
\begin{table}[]
    \centering
    \small
    \begin{tabular}{c|c c  }
    \toprule 
        Model  &  Span F1 & Head F1\\
        \midrule 
       Seq   &  40.5 & 48.0   \\
       Two-step & 41.8 & 49.7  \\
       
       \ours & \textbf{48.64} & \textbf{57.32} \\
       \bottomrule 
       
    \end{tabular}
    \caption{Supervised argument extraction results (\%) on RAMS test set. Both the \textbf{Seq} and \textbf{Two-step} model use type-constrained decoding to improve performance. 
    The official scorer was used to compute results.
    }
    \label{tab:RAMS-AE}
\end{table}

\begin{table}[t]
    \centering
    \small
    \resizebox{\hsize}{!}{
    \begin{tabular}{c|c  c c c }
    \toprule 
    Model & \multicolumn{2}{c}{Arg Identification} & \multicolumn{2}{c}{Arg Classification}\\
    \midrule
          &  Head F1 & Coref F1 &Head F1 & Coref F1 \\
        \midrule 
        BERT-CRF & 69.83 & 72.24  & 54.48 & 56.72 \\
        BERT-QA  &61.05 &  64.59 & 56.16 &  59.36  \\
        BERT-QA-Doc & 39.15 & 51.25 & 34.77 & 45.96 \\
      \ours & \textbf{71.75}  & \textbf{72.29} & \textbf{64.57} & \textbf{65.11} \\
      \bottomrule 
    \end{tabular}}
    \caption{Argument extraction results (\%) on \textsc{WikiEvents} test set. }
    \label{tab:KAIROS-AE}
\end{table}

\subsection{Document-Level Informative Argument Extraction}
We test on \textsc{WikiEvents} using the informative argument as the training data and also compare with the BERT-CRF and BERT-QA baselines. 
Results are shown in Table \ref{tab:KAIROS-info-arg}.

\begin{table}[t]
    \centering
    \small
    \resizebox{\hsize}{!}{
    \begin{tabular}{c|  c    c  c c }
    \toprule 
    Model & \multicolumn{2}{c }{Arg-I} & \multicolumn{2}{c}{Arg-C} \\
      & Head F1 & Coref F1  & Head F1 & Coref F1 \\
        \midrule 
        BERT-CRF & 52.71 & 58.12 & 43.29 &  47.70 \\
        BERT-QA & 46.88 & 49.89 & 43.44 & 46.45 \\
        BERT-QA-Doc & 26.29 & 31.54 & 24.46 & 29.26 \\
      \ours & \textbf{56.10} & \textbf{62.48} & \textbf{51.03} & \textbf{57.04} \\ 
      \bottomrule 
    \end{tabular}}
    \caption{Informative argument extraction results on \textsc{WikiEvents} test set. }
    \label{tab:KAIROS-info-arg}
\end{table}

Comparing the results in Tables \ref{tab:KAIROS-AE} and \ref{tab:KAIROS-info-arg}, we have the following findings:
\begin{enumerate}
    \item Informative argument extraction is a much more difficult task compared to nearest argument extraction. This is exemplified by the large performance gap for all models.
    \item While CRF models are good at identifying spans, the  performance is hindered by classification. The arguments follow a long tail distribution and since CRF models learn each argument tag separately, it cannot leverage the similarity between argument roles to improve the performance on rarely seen roles. 
    \item QA models, on the other hand, suffer from poor argument identification. When the QA model produces multiple answers for the same role, these answer spans are often close to each other or overlap. We show a concrete example in the qualitative analysis. 
    \item Directly applying the BERT-QA model to document level does not work. The QA model gets easily distracted by the additional context and does not know which event to focus on. We think that this is not a fundamental limitation of the QA approach, but a sign that repurposing QA models for document-level event extraction needs more investigation.
\end{enumerate}

\begin{table*}[]
    \centering
    \small 
    \resizebox{\hsize}{!}{
    \begin{tabular}{l |l  c c c }
    \toprule 
    Context     &  Role & Ours & BERT-CRF & BERT-QA \\
    \midrule 
     \multirow{8}{23em}{
     \textbf{I} have been \underline{in touch} (E1:Contact.Contact.Correspondence) with the NDS \textbf{official} in the province and they told me that over 100 \textbf{members} of the NDS were \underline{killed}(E2:Life.Die)  in the big explosion , " the former provincial official said .  
     Sharif Hotak , a member of the provincial council in Maidan Wardak said \textbf{he} \underline{saw} 
     (E3:Cognitive.IdentifyCategorize) \textbf{bodies}  of 35 Afghan forces in the \textbf{hospital}."} 
     & E1: Participant &  I & I  & NDS \\
     & &  NDS official & official & I  \\
     & & & & official \\
     & & & & NDS official \\
     & E2: Victim & members & members & members\\
     & E3: Identifier &  he & N/A& N/A \\
     & E3: IdentifiedObject &  bodies & N/A& N/A \\
     & E3: Place & hospital & N/A& N/A \\
     \bottomrule 
    \end{tabular}}
    \caption{An example of document-level argument extraction task on \textsc{WikiEvents}. This excerpt contains 3 events: Contact, Die and IdentifyCategorize. For the Contact event E1, BERT-QA over-generates answers for the participant span. For the Die event, all three models can correctly extract the Victim argument. For the IdentifyCategorize event which is relatively rare, only our model can successfully extract all arguments.
    }
    \label{tab:arg-ex}
\end{table*}

\begin{figure*}[t]
    \centering
    \includegraphics[width=.9\linewidth]{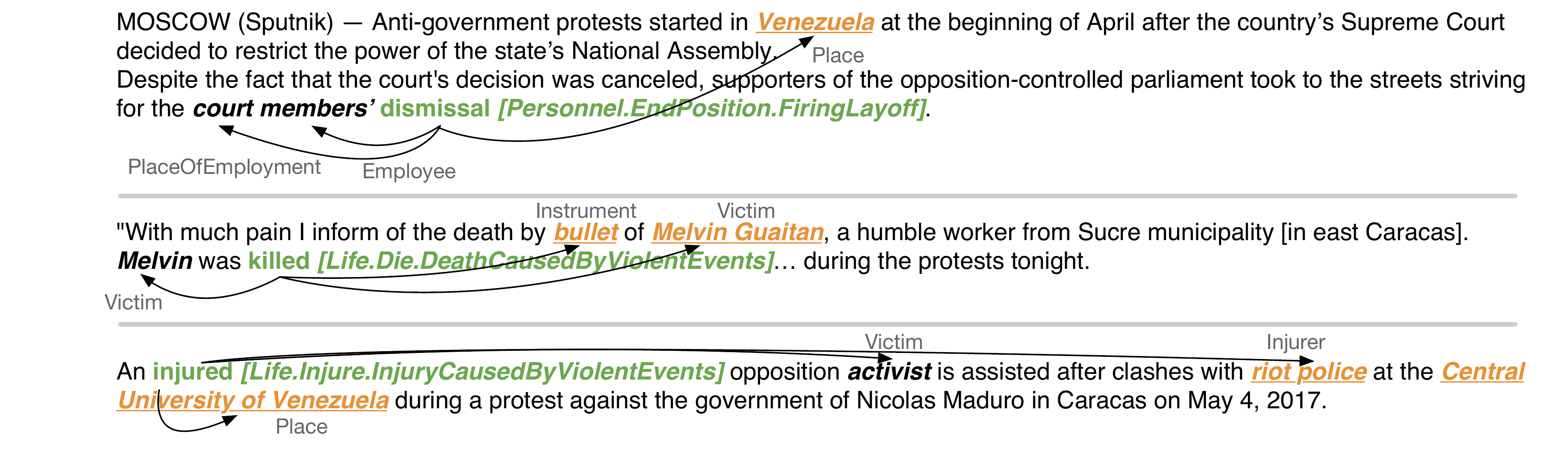}
    \caption{An example of our model's argument extraction output on the SM-KBP dataset. Arguments highlighted in orange are newly found by our model compared to the baseline system OneIE~\cite{Lin2020OneIE}. }
    \label{fig:aida}
\end{figure*}

\begin{figure}
    \centering
    \includegraphics[width=\linewidth]{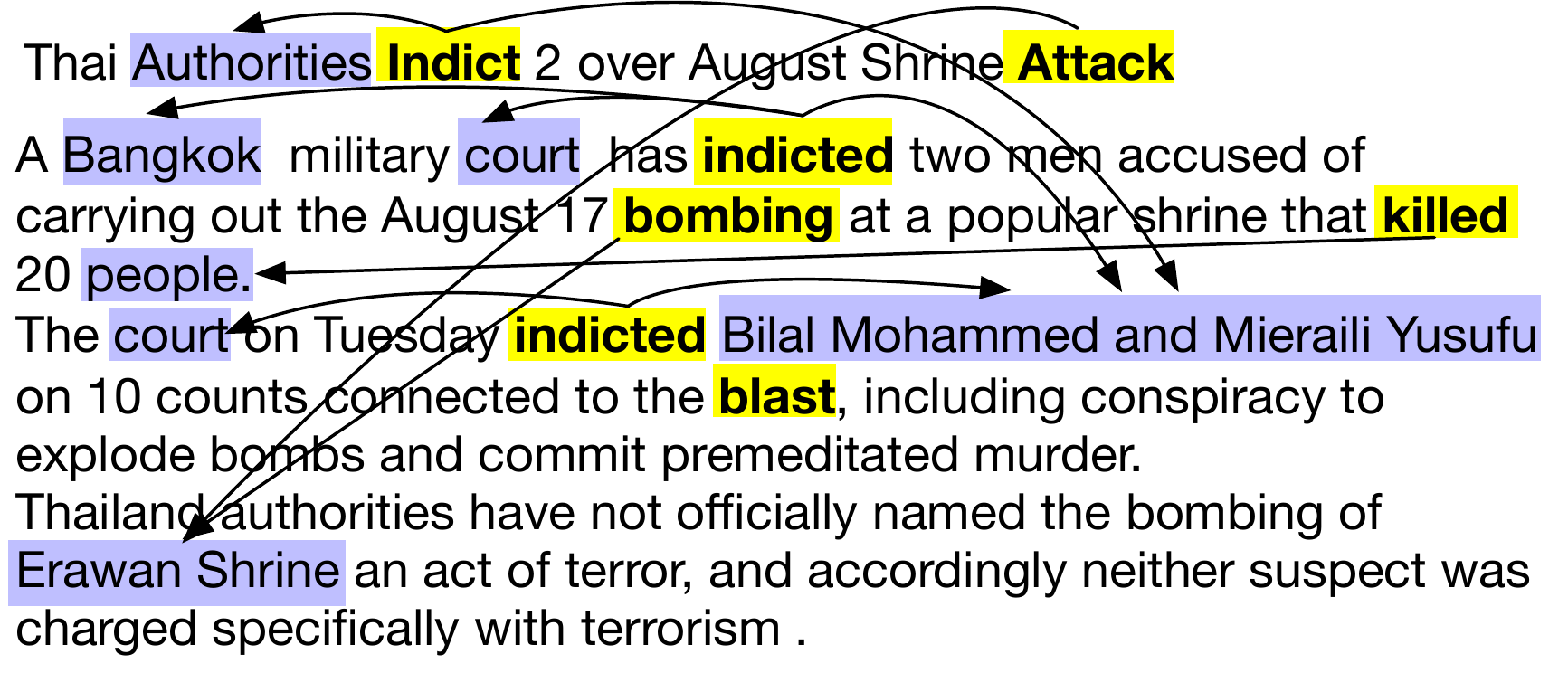}
    \caption{An example of our model's prediction of informative arguments. Only arguments for the boldfaced event triggers are shown. Notably, our model can correctly identify ``Bilal Mohammed and Mieralli Yusufu" as the as the Defendant for all the ChargeIndict events and ``Erawan Shrine" as the place of attack.}
    \label{fig:info-ex}
\end{figure}

\subsection{Zero-Shot Event Extraction}
We show results for the zero-shot transfer setting in Table \ref{tab:zero-shot}.
Since the baseline BERT-CRF model~\cite{shi2019simpleBERTCRF} cannot handle new labels directly, we exclude it from comparison.
In addition to BERT-QA, we also replace our \textsc{TapKey} trigger extraction model with a Prototype Network\cite{Snell2017PrototypicalNF}\footnote{When 0 event types are seen, we set the transformation function in the Prototype Network to be an identity function; in other cases, we use a two layer MLP.}. %
We replace the prototypes with the class vectors to enable zero-shot learning.  Complete results for trigger extraction are included in Appendix D.

The performance of BERT-QA is greatly limited by the trigger identification step. Both the Prototype network and our model \textsc{TapKey} can leverage the keyword information to assist transfer. 
Remarkably, \textsc{TapKey} has only 3 points drop in F1 using only 30\% of the training data compared to the full set. 
The argument extraction component is more sensitive to the reduction in training data, but still performs relatively well. 
We notice that when a template is completely new, the model might alter the template structure during generation. 

\begin{table*}[t]
    \centering
    \small 
    \begin{tabular}{c|c| c c c c   }
          \toprule 
      \# Seen Event Types &  Model  & TI F1 &  TC F1 & AI Head F1 &  AC Head F1  \\
      \midrule 
       \multirow{2}{*}{0} 
       & Prototype &  3.19 & 2.22  & - & -  \\
      & \textsc{TapKey} &  55.19 &  52.10 &- & -\\
      \midrule 
       \multirow{3}{*}{10 most frequent} 
      & BERT-QA &  57.06 &  53.83 & 39.37  & 38.27  \\
         & Prototype + BART-Gen  & 69.48  & 66.06 & 46.08 & 41.93  \\
       & \textsc{TapKey} + BART-Gen &  72.31  & 69.23  & 48.18 &  44.19 \\
      \midrule 
       \multirow{3}{*}{1 per general type} 
      & BERT-QA & 27.56 & 25.32 &24.87  & 24.29 \\
      & Prototype + BART-Gen & 68.29 & 65.85 & 39.51 &  34.63  \\
      & \textsc{TapKey} + BART-Gen  & 72.23 & 68.55 & 39.80  & 35.11 \\
      \midrule 
      \multirow{1}{*}{Full} 
      &  \textsc{TapKey} + BART-Gen & 74.36 & 71.13 & 55.22 & 53.71 \\ 
         \bottomrule 
    \end{tabular}
    \caption{Zero-shot event extraction results (\%) on \textsc{ACE}. ``10 most frequent event types" corresponds to the \textbf{Freq} data split and ``1 per general type" corresponds to the \textbf{Ontology} data split. Fully supervised results are provided for reference. }
    \label{tab:zero-shot}
    \vspace{-0.2cm}
\end{table*}

\subsection{Qualitative Analysis}
We show a comparison of our model's extractions with baselines in Table \ref{tab:arg-ex} for the argument extraction task on \textsc{WikiEvents}.
Our model is able to effectively capture all arguments, while the CRF model struggles with rare event types and the QA model is hindered by over-generation.

An example of informative argument extraction from our model is displayed in Figure \ref{fig:info-ex}. 
Our model is able to choose the informative mentions of the Defendant of indiction and Place of attack even when the trigger is a few sentences away.

We also applied our model as part of a pipeline multimedia multilingual knowledge extraction system~\cite{GAIASystem2020} for the NIST streaming multimedia knowledge base population task (SM-KBP2020)\footnote{\url{https://tac.nist.gov/2020/KBP/SM-KBP/}}. Our model was able to discover 53\% new arguments compared to the original system, especially for those that were further away from the event trigger. The overall system achieved top 1 performance. We show some examples in Figure \ref{fig:aida}.

\subsection{Remaining Challenges}
\noindent \textbf{Ontological Constraints} 
Some of the roles are mutually exclusive, such as the Origin/Destination in the Transport event and the Recipient/Yielder in the Surrender event.
In the following example, ``Japan" was extracted as both part of the Recipient and the Yielder of the Surrender event: 
\textit{``If South Korea drifts into the orbit of the \textbf{US and Japan}, China's influence on the Korean peninsula could be badly compromised." At a military parade in Beijing to mark the 70th anniversary of the \underline{surrender} of \textbf{Japan} last September, ..."}. 
Such constraints might be incorporated into the decoding process of the model.

\noindent \textbf{Commonsense Knowledge}
In the following instance with implicit arguments: \textit{``Whether the \textbf{U.S. } extradites \textbf{Gulen}  or not this will be a political decision, ”Bozdag said.“ If he is not \underline{extradited}, \textbf{Turkey} will have been sacrificed for a terrorist.” A recent opinion poll showed two thirds of Turks agree with their president that Gulen was behind the \textbf{coup plot.}"}, our model mistakenly labels ``U.S." as the Destination of the extradition and ``Turkey" as the Source even though the Extraditer is correctly identified as ``U.S.". Commonsense knowledge such as ``The extraditer, if being a country, is usually is same as the source of extradition" would be helpful to fix this error.

%% file: related.tex
\section{Related Work}

\subsection{Document-Level Event Extraction}
Document-level event extraction can be traced back role filling tasks from the MUC conferences~\cite{Grishman1996MUC6} that required retrieving participating entities and attribute values for specific scenarios. The KBP slot filling challenge\footnote{\url{https://tac.nist.gov/2017/KBP/index.html}} is akin to this task, but centered upon entities. 

In general, document-level argument extraction is an under-explored topic, mainly due to the lack of datasets.
There have been a few datasets published specifically for implicit semantic role labeling, such as the SemEval 2010 Task 10~\cite{Ruppenhofer2010SemEval2010T1}, the Beyond NomBank dataset~\cite{Gerber2010BeyondNA} and ON5V~\cite{Moor2013PredicatespecificAF}. However, these datasets were small in size and only covered a small set of carefully selected predicates. 
Recently, \cite{Ebner2020RAMS} published the RAMS dataset, which contains annotation for cross-sentence implicit arguments covering a wide range of event types.
Albeit, this dataset only annotates one event per document, motivating us to create a new benchmark with complete event and coreference annotation.

The GRIT model~\cite{Du2020GRITGR} is a generative model designed for the MUC task which can be seen as filling in predefined tables.  
In comparison, we treat the template (for example "<arg1> attacked <arg2> using <arg3> at <arg4> place") as part of the model input along with the document context. This allows us to share model parameters across all event types and enables zero-shot transfer to new event types.

\subsection{Zero-shot Event Extraction}
Early attempts at zero-shot or few-shot event extraction rely on preprocessing such as semantic role labeling(SRL)~\cite{Peng2016EventDA} or abstract meaning representation (AMR)~\cite{Huang2018ZeroShotTL} to detect trigger mentions and argument mentions before performing classification on the detected spans.

Another line of work only examines the subtask of trigger detection, essentially reducing the task to few-shot classification. Both \cite{Lai2020ExtensivelyMF} and \cite{Deng2020MetaLearningWD} extend upon the prototype network model~\cite{Snell2017PrototypicalNF} for classification.

Recent work on zero-shot event extraction has posed the problem as question answering~\cite{Chen2019ReadingTheManual, Du2020EventQA, Feng2020ProbingAF} with different ways of designing the questions.

%% file: conclusion.tex
\section{Conclusion \& Future Work}
In this paper, we advocate document-level event extraction and propose the first document-level neural event argument extraction model. 
We also release the first document-level event extraction benchmark dataset~\textsc{WikiEvents} with complete event and coreference annotation.
On both the conventional argument extraction task and the new informative argument extraction task, our proposed model surpasses CRF-based and QA-based baselines by a wide margin.
Additionally, we demonstrate the portability of our model by applying it to the zero-shot setting. 
Going forward, we would like to incorporate more ontological knowledge to produce more accurate extractions.